\documentclass[runningheads]{llncs}
\usepackage{graphicx}
\usepackage{amsmath,amssymb} 
\usepackage{color}
\usepackage{csquotes}
\usepackage{cite}

\usepackage[small]{caption}
\usepackage{booktabs}       
\usepackage{mathtools}
\usepackage{xspace}
\usepackage[caption=false,farskip=0pt,justification=centerlast,font=scriptsize,labelformat=simple]{subfig}

\begin{document}
\pagestyle{headings}
\mainmatter
\def\ECCV16SubNumber{***}  

\title{Instance- and Category-level 6D Object Pose Estimation} 



\author{Caner Sahin, Guillermo Garcia-Hernando, Juil Sock, \\and Tae-Kyun Kim}
\institute{ICVL, Imperial College London}

\maketitle

\begin{abstract}
6D object pose estimation is an important task that determines the 3D position and 3D rotation of an object in camera-centred coordinates. By utilizing such a task, one can propose promising solutions for various problems related to scene understanding, augmented reality, control and navigation of robotics. Recent developments on visual depth sensors and low-cost availability of depth data significantly facilitate object pose estimation. Using depth information from RGB-D sensors, substantial progress has been made in the last decade by the methods addressing the challenges such as viewpoint variability, occlusion and clutter, and similar looking distractors. Particularly, with the recent advent of convolutional neural networks, RGB-only based solutions have been presented. However, improved results have only been reported for recovering the pose of known instances, i.e., for the instance-level object pose estimation tasks. More recently, state-of-the-art approaches target to solve object pose estimation problem at the level of categories, recovering the 6D pose of unknown instances. To this end, they address the challenges of the category-level tasks such as distribution shift among source and target domains, high intra-class variations, and shape discrepancies between objects.
\end{abstract}
\section{Introduction}
6D object pose estimation is an important problem in the realm of computer vision that determines the 3D position and 3D orientation of an object in camera-centered coordinates \cite{b28}. It has extensively been studied in the past decade as it is of great importance to many rapidly evolving technological areas, such as robotics and augmented reality. Particularly, increasing ubiquity of Kinect-like RGB-D sensors and low-cost availability of depth data facilitate object pose estimation scenarios related to the above-mentioned areas.\\
\indent Robotic manipulators that pick and place the goods from conveyors, shelves, pallets, etc., facilitate several processes comprised within logistics systems, \textit{e.g.}, warehousing, material handling, packaging \cite{76, 77, 78, 79}. Amazon Picking Challenge (APC) \cite{a16} is an important example demonstrating the promising role of robotic manipulation for the facilitation of such processes. APC integrates many tasks, such as mapping, motion planning, grasping, object manipulation, with the goal of autonomously moving items by robotic systems from a warehouse shelf into a tote. Regarding the automated handling of items by robots, accurate object detection and 6D pose estimation is an important task that when successfully performed improves the autonomy of the manipulation. Household robotics is another field where the ability of recognizing objects and accurately estimating their poses is a key element. This capability is needed for such robots, since they should be able to navigate in unconstrained human environments calculating grasping and avoidance strategies. In this scenario, unlike industrial applications, the workspace is completely unknown, and thus making indispensable the existence of 6D pose estimators, which are highly robust to changing, dynamic environments. Aerial images are required to automatically analysed to recognize abnormal behaviors in target terrains \cite{b38}. Unmanned aerial vehicles function surveillance and reconnaissance in order to ensure high level security detecting and estimating 6D poses of interested objects \cite{80, 81, 82}. Virtual Reality (VR) and Augmented Reality (AR) systems need to know accurate positions, poses, and geometric relations of objects so as to locate virtual objects in the real world \cite{83, 84, 85}.\\
\section{Problem Formulation}
\label{ch1_problem_form}
\indent 6D object pose estimation is addressed in the literature at the level of both instances and categories. Instance-level 6D object pose estimation tasks require the same statistical distribution both for source data from which a classifier is learnt and for target data on which the classifiers will be tested. Hence, instance-based methods estimate 6D poses of \textit{seen} objects, mainly targeting to report improved results overcoming instances' challenges, such as viewpoint variability, occlusion, clutter, and similar-looking distractors. However, instance-based methods cannot easily be generalized for category-level 6D object pose estimation tasks, which inherently involve the challenges such as distribution shift among source and target domains, high intra-class variations, and shape discrepancies between objects.\\
\index{instance-based 6D object pose estimation}
\indent Formally 6D object pose estimation can be defined as a prediction problem, and we formulate this problem at the level of \enquote{instances} as follows: Given an RGB-D image $I$ where an instance $S$ of the interested object $O$ exists, we cast 6D object pose estimation as a joint probability estimation problem, and formulate it as given below:
\begin{equation}
{(\mathbf{x}, \mathbf{\theta})}^* = \arg \max_{\mathbf{x}, \mathbf{\theta}}  p(\mathbf{x}, \mathbf{\theta} \vert I, S)
\label{eq1}
\end{equation}
where $\mathbf{x} = (x, y, z)$ is the 3D translation and $\mathbf{\theta} = (r, p, y)$ is the 3D rotation of the instance $S$. $(r, p, y)$ depicts the Euler angles, roll, pitch, and yaw, respectively. According to Eq. \ref{eq1}, methods for 6D object pose estimation problem target to maximize the joint posterior density of the 3D translation $\mathbf{x}$ and 3D rotation $\mathbf{\theta}$. This formulation assumes that there only exists one instance of the interested object in the RGB-D image $I$, and hence, producing the pair of pose parameters ($\mathbf{x}$, $\mathbf{\theta}$), which is of the instance $S$. Note that, this existence is apriori known by any 6D object pose estimation method.\\
\index{category-based 6D object pose estimation}
\indent In case the image $I$ involves multiple instances $\mathcal{S} = \{ S_i \vert i = 1, ... ,n \}$ of the object of interest, the formulation of the problem takes the following form:
\begin{equation}
{(\mathbf{x}_i, \mathbf{\theta}_i)}^* = \arg \max_{\mathbf{x}_i, \mathbf{\theta}_i}  p(\mathbf{x}_i, \mathbf{\theta}_i \vert I, \mathcal{S}), \quad i = 1, ..., n.
\label{ch1_eq1}
\end{equation}
The number of instances $n$ is apriori known by the methods. Given an instance $C$ of an interested category $c$, the 6D object pose estimation problem is formulated at the level of categories transforming Eq. \ref{eq1} into the following form:
\begin{equation}
{(\mathbf{x}, \mathbf{\theta})}^* = \arg \max_{\mathbf{x}, \mathbf{\theta}}  p(\mathbf{x}, \mathbf{\theta} \vert I, C, c).
\label{eq3}
\end{equation}
Equation \ref{eq3} assumes that there is only one instance of the interested category in the RGB-D image $I$ (apriori known by any 6D object pose estimation method), and hence, producing the pair of pose parameters ($\mathbf{x}$, $\mathbf{\theta}$), which is of the instance $C$. In case the image $I$ involves multiple instances $\mathcal{C} = \{ C_i \vert i = 1, ... ,n \}$ of the category of interest, Eq. \ref{ch1_eq1} takes the following form, where the number of instances $n$ is apriori known:
\begin{equation}
{(\mathbf{x}_i, \mathbf{\theta}_i)}^* = \arg \max_{\mathbf{x}_i, \mathbf{\theta}_i}  p(\mathbf{x}_i, \mathbf{\theta}_i \vert I, \mathcal{C}, c), \quad i = 1, ..., n.
\label{ch1_eq2}
\end{equation}
\section{Challenges of the Problem}
\label{challenges}
\index{challenges}
Any method engineered for 6D object pose estimation has to cope with the challenges of the problem, in order to robustly work in a generalized fashion. We categorize these challenges according to the level at which they are observed: challenges of instances and challenges of categories. Note that instances' challenges can also be observed at the level of categories, but not the other way round.\\

\noindent \textbf{Challenges of Instances:} The challenges mainly encountered at the level of instances are viewpoint variability, texture-less objects, occlusion, clutter, and similar-looking distractors.\\
\index{challenges of instances!viewpoint variability}
\indent \textit{\textbf{Viewpoint variability.}} Testing scenes, where target objects are located, can be sampled to produce sequences that are widely distributed in the pose space by $[0 - 360]$ degree, $[-180 - 180]$ degree, $[-180 - 180]$ degree in the roll, pitch, and yaw angles, respectively. As the pose space gets wider, the amount of data required for training a 6D estimator increases, in order to capture reasonable viewpoint coverage of the target object.\\
\index{challenges of instances!texture-less objects}
\indent \textit{\textbf{Texture-less objects.}} Texture is an important information for RGB cameras, which can capture and represent a scene by 3 basic colors (channels): red, green and blue. An object of interest can easily be distinguished from background or any other instances available in the scene, in case it is sufficiently textured. This is mainly because of that texture on the surface allows to define discriminative features to represent the interested object. However, when objects are texture-less, this discriminative property disappears, and thus making methods strongly dependent on depth channel in order to estimate 6D poses of objects. \\
\index{challenges of instances!occlusion}
\indent \textit{\textbf{Occlusion.}} As being one of the most common challenges observed in 6D object pose estimation, occlusion occurs when an object of interest is partly or completely blocked by other objects existing in the scene. Naive occlusion is handled by either modelling it during an off-line training phase or engineering a part-based approach that infers the 6D pose of the object of interest from its unoccluded (occlusion-free) parts. However, the existence of severe occlusion gives rise false positive estimations, degrading methods' performance.\\
\index{challenges of instances!clutter}
\indent \textit{\textbf{Clutter.}} Clutter is a challenge mainly associated with complicated backgrounds of images, in which existing objects of interest even cannot be detected by naked eye. Several methods handle this challenge training the algorithms with cluttered background images. However, utilizing background images alleviates the generalization capability of methods, making those data-dependent.\\
\index{challenges of instances!similar looking distractors}
\indent \textit{\textbf{Similar-Looking Distractors.}} Similar-looking distractors along with similar looking object classes is one of the biggest challenges observed in 6D object pose recovery. In case the similarity is in depth channel, 6D pose estimators are strongly confused because of the lack of discriminative selection of shape features. The lacking in shape is compensated by RGB in case there is no color similarity.\\

\noindent \textbf{Challenges of Categories:} The challenges mainly encountered at category-level 6D object pose estimation are intra-class variation and distribution shift.\\
\index{challenges of categories!intra-class variation}
\indent \textit{\textbf{Intra-class variation.}} Despite the fact that instances from the same category typically have similar physical properties, they are not exactly the same. While texture and color variations are seen in RGB channel, geometry and shape discrepancies are observed in depth channel. Geometric dissimilarities are related to scale and dimensions of the instances, and shape-wise, they appear different in case they physically have extra parts out of the common ones. Category-level 6D object pose estimators handle intra-class variation during training using the data that are of the instances belonging to the source domain.\\
\index{challenges of categories!distribution shift}
\indent \textit{\textbf{Distribution shift.}} Any 6D pose estimator working at the level of categories is tested on the instances in target domain. Since the objects in the target domain are different than that are of the source domain, there is a shift between the marginal probability distributions of these two domains. Additionally, this distribution shift itself also changes as the instances in the target domain are unseen to the 6D pose estimator.
\section{Methods}
\label{baselines}
\index{instance-based methods}
State-of-the-art baselines for 6D object pose estimation address the challenges studied in Sect. \ref{challenges}, however, the architectures used differ between the baselines. In this section, we analyse 6D object pose estimators architecture-wise.\\

\noindent \textbf{Instance-based Methods:} This family involves template-based, point-to-point, conventional learning-based, and deep learning methods.\\
\index{instance-based methods!template-based}
\indent \textit{\textbf{Template-based.}} Template-based methods involve an off-line template generation phase. Using the 3D model $M$ of an interested object $O$, a set of RGB-D templates are synthetically rendered from different camera viewpoints. Each pair of RGB and depth images is annotated with 6D pose and is represented with feature descriptors. The 3D model $M$ can either be a CAD or a reconstructed model. A template-based method takes an RGB-D image $I$ as input on which it runs a sliding window during an on-line test phase. Each of the windows is represented with feature descriptors and is compared with the templates stored in a memory. The distances between each window and the template set are computed. The 6D pose of a template is assigned to the window that has the closest distance with that template \cite{63, 32, x1, 34}.\\
\index{instance-based methods!point-to-point}
\indent \textit{\textbf{Point-to-point.}} These methods simultaneously estimate object location and pose by firstly establishing for each scene point a spatial correspondence to a model point, and then rotationally aligning the scene to model point cloud. To this end, point pair features (PPF) are used along with a voting scheme. Both models and scenes are represented with point pair features (PPF). During an on-line stage, a set of point pair features are computed from the input depth image $I_D$. Created point pairs are compared with the ones stored in the global model representation. This comparison is employed in feature space, and a set of potential matches, and the corresponding 6D pose are resulted \cite{64, 66, x2, x3, x4, 76}.\\
\index{instance-based methods!conventional learning-based}
\indent \textit{\textbf{Conventional Learning-based.}} Conventional learning-based methods are mainly learnt based on two approaches during an off-line step: i) holistic and ii) part-based. In holistic learning, the process of generating holistic training data is the same as the template generation phase of \enquote{template-based methods}. In part-based learning, a set of patches is extracted from each pair of RGB and depth images available in the holistic training data. Each of extracted patches is annotated with 6D pose and is represented with features. The holistic and part-based training data are separately used to train a regressor, which can be a random forest, nearest-neighbor classifier, or an SVM. In turn, during an on-line inference stage, an RGB-D image $I$ is taken as input by a conventional learning-based method. If the method is holistic, bounding boxes are extracted from $I$ running a sliding window through, and they are fed into a holistic regressor \cite{73, 30}. In case the method is based on parts, extracted patches are sent to a part-based regressor \cite{20, 31, 36, 21, 19, 67, 71}. Both types of regressors output 6D pose parameters $(\mathbf{x}, \mathbf{\theta})$. Several conventional learning-based methods employ a final pose refinement step in order to further cure the 6D pose \cite{36}. As ICP-like algorithms are used to further refine the pose, classifier/regressor itself is also engineered so that this refinement employed architecture-wise \cite{20, 19, 67}.\\
\index{instance-based methods!deep learning}
\indent \textit{\textbf{Deep learning.}} Current paradigm in the community is to learn deep discriminative feature representations \cite{x14}. Wohlhart et al. \cite{38} utilize a CNN structure to learn discriminative descriptors and then pass the learnt descriptors to a Nearest Neighbor classifier in order to find the closest object pose. Although promising, this method has one main limitation, which is the requirement of background images during training along with the ones holistic foreground, thus making its performance dataset-specific. The studies in \cite{18, 33} learn deep representation of parts in an unsupervised fashion only from foreground images using auto-encoder architectures. The features extracted in the course of the test are fed into a Hough forest in \cite{18}, and into a codebook of pre-computed synthetic local object patches in \cite{33} in order to hypothesise object 6D pose. While \cite{38} focuses on learning feature embeddings based on metric learning with triplet comparisons, Balntas et al. \cite{74} further examine the effects of using object poses as guidance to learning robust features for 3D object pose estimation in order to handle symmetry issue.\\
\indent More recent methods adopt CNNs for 6D pose estimation, taking RGB images as inputs \cite{72}. BB8 \cite{69} and Tekin et al. \cite{70} perform corner-point regression followed by PnP for 6D pose estimation. Typically employed is a computationally expensive post processing step such as iterative closest point (ICP) or a verification network \cite{68}.\\

\index{category-based methods}
\noindent \textbf{Category-based Methods:} There have been a large number of methods addressing object detection and pose estimation at the level of categories, however, none of those methods are engineered to estimate the full 6D poses of the instances of a given category, as formulized in Sect. \ref{ch1_problem_form}, out of the architecture presented in \cite{75}. Our classification for category-level methods are based on the dimension concerned.\\
\index{category-based methods!2D}
\indent \textit{\textbf{2D.}} One line of the methods is based on visual perception in RGB channel. Deformable part models \cite{a49, a50, a53} are designed to work in RGB, detecting objects of the interested category in 2D. More recent paradigm is to learn generic feature representations on which fine-tuning will later be applied. CNN based approaches \cite{a51} have been developed for this purpose, however, they require large-scale annotated images to provide the generalization on feature representations \cite{a45}. Since these approaches work in the context of RGB modality, the success of such methods is limited to coarse/discrete solutions in 2D. Several studies exploit 3D geometry fusing depth channel with RGB \cite{27, 28}. They mainly use CNN architectures in order to learn representations, which are then fed into SVM classifiers. Even though performance improvement is observed, they are not generalised well to go beyond 2D applications. \\
\index{category-based methods!3D}
\indent \textit{\textbf{3D.}} Methods engineered for 3D object detection focus on finding the bounding volume of objects and do not predict the 6D pose of the objects \cite{x5, x6, x7, x8}. While \cite{x9} directly detect objects in 3D space taking 3D volumetric data as input, the studies in \cite{x10, x11, x12} first produce 2D object proposals in 2D image and then project the proposal into 3D space to further refine the final 3D bounding box location.\\
\index{category-based methods!4D}
\indent \textit{\textbf{4D.}} SVM-based Sliding Shapes (SS) \cite{2} method detects objects in the context of depth modality naturally tackling the variations of texture, illumination, and viewpoint. The detection performance of this method is further improved in Deep Sliding Shapes (Deep SS) \cite{1}, where more powerful representations encoding geometric shapes are learned in ConvNets. These two methods run sliding windows in the 3D space mainly concerning 3D object detection of bounding boxes aligned around the gravity direction, rather than full 6D pose estimation. The system in \cite{4}, inspired by \cite{2}, estimates detected and segmented objects' rotation around the gravity axis using a CNN. The system is the combination of individual detection/segmentation and pose estimation frameworks. The ways the methods above \cite{2, 1, 4} address the challenges of categories are relatively naive. Both SS and the method in \cite{4} rely on the availability of large scale 3D models in order to cover the shape variance of objects in the real world. Deep SS performs slightly better against the categories' challenges, however, its effort is limited to the capability of ConvNets.\\
\index{category-based methods!6D}
\indent \textit{\textbf{6D.}} The study in \cite{75} presents \enquote{Intrinsic Structure Adaptors (ISA)}, a part-based random forest architecture, for full 6D object pose estimation at the level of categories in depth images. To this end, \textit{3D skeleton structures} are derived as shape-invariant features, and are used as privileged information during the training phase of the architecture.
\section{Datasets}
\label{datasets}
Every dataset used in this study is composed of several object classes, for each of which a set of RGB-D test images are provided with ground truth object poses.\\

\noindent \textbf{Datasets of Instances:} The collected datasets of instances mainly differ from the point of the challenges that they involve (see Table \ref{tab_1}).\\
\index{challenges of instances!clutter}
\index{challenges of instances!viewpoint variability}
\indent \textit{\textbf{Viewpoint (VP) + Clutter (C).}} Every dataset involves the test scenes in which objects of interest are located at \textit{varying viewpoints} and \textit{cluttered backgrounds}.\\
\index{challenges of instances!texture-less objects}
\indent \textit{\textbf{VP + C + Texture-less (TL).}} Test scenes in the LINEMOD \cite{32} dataset involve \textit{texture-less} objects at varying viewpoints with cluttered backgrounds. There are 15 objects, for each of which more than $1100$ real images are recorded. The sequences provide views from $0$ - $360$ degree around the object, $0$ - $90$ degree tilt rotation, $\mp 45$ degree in-plane rotation, and $650$ mm - $1150$ mm object distance.\\
\index{challenges of instances!occlusion}
\indent \textit{\textbf{VP + C + TL + Occlusion (O) + Multiple Instance (MI).}} Occlusion is one of the main challenges that makes the datasets more difficult for the task of object detection and 6D pose estimation. In addition to close and far range 2D and 3D clutter, testing sequences of the Multiple-Instance (MULT-I) dataset \cite{20} contain \textit{foreground occlusions} and \textit{multiple object instances}. In total, there are approximately $2000$ real images of $6$ different objects, which are located at the range of $600$ mm - $1200$ mm. The testing images are sampled to produce sequences that are uniformly distributed in the pose space by $[0^\circ - 360^\circ ]$, $[-80^\circ - 80^\circ ]$, and $[-70^\circ - 70^\circ ]$ in the yaw, roll, and pitch angles, respectively.\\
\indent \textit{\textbf{VP + C + TL + Severe Occlusion (SO).}} Occlusion, clutter, texture-less objects, and change in viewpoint are the most well-known challenges that could successfully be dealt with the state-of-the-art 6D object detectors. However, \textit{heavy existence} of these challenges severely degrades the performance of 6D object detectors. Occlusion (OCC) dataset \cite{31} is one of the most difficult datasets in which one can observe up to  $70-80 \%$ occluded objects. OCC includes the extended ground truth annotations of LINEMOD: in each test scene of the LINEMOD \cite{32} dataset, various objects are present, but only ground truth poses for one object are given. Brachmann et al. \cite{31} form OCC considering the images of one scene (benchvise) and annotating the poses of 8 additional objects.\\
\indent \textit{\textbf{VP + SC + SO + MI + Bin Picking (BP).}} In \textit{bin-picking} scenarios, multiple instances of the objects of interest are arbitrarily stocked in a bin, and hence, the objects are inherently subjected to severe occlusion and severe clutter. Bin-Picking (BIN-P) dataset \cite{18} is created to reflect such challenges found in industrial settings. It includes $183$ test images of $2$ textured objects under varying viewpoints.\\
\index{challenges of instances!similar looking distractors}
\indent \textit{\textbf{VP + C + TL + O + MI + Similar Looking Distractors (SLD).}} \textit{Similar-looking distractor(s)} along with similar looking object classes involved in the datasets strongly confuse recognition systems causing a lack of discriminative selection of shape features. Unlike the above-mentioned datasets and their corresponding challenges, the T-LESS \cite{61} dataset particularly focuses on this problem. The RGB-D images of the objects located on a table are captured at different viewpoints covering $360$ degrees rotation, and various object arrangements generate occlusion. Out-of-training objects, similar looking distractors (planar surfaces), and similar looking objects cause $6$ DoF methods to produce many false positives, particularly affecting the depth modality features. T-LESS has $30$ texture-less industry-relevant objects, and $20$ different test scenes, each of which consists of $504$ test images.\\
\begin{table*}[t]
\caption{Datasets collected: each dataset shows different characteristics mainly from the challenge point of view (VP: viewpoint, O: occlusion, C: clutter, SO: severe occlusion, SC: severe clutter, MI: multiple instance, SLD: similar looking distractors, BP: bin picking).}
\vspace{0.8em}
\centering
\setlength\tabcolsep{6pt}
\begin{minipage}{\textwidth}
\centering
\resizebox{0.99\columnwidth}{!}{
\begin{tabular}{ l c c c c c}
    \toprule
    \textbf{Dataset} & Challenge & \# Obj. Classes & Modality & \# Total Frame & Obj. Dist. [mm]\\ 
    \midrule
    LINEMOD          & VP + C + TL               &$15$ &RGB-D &15770 &600-1200 \\ 
    MULT-I           & VP + C + TL + O + MI      &$6$  &RGB-D &2067  &600-1200\\
    OCC              & VP + C + TL + SO          &$8$  &RGB-D &9209  &600-1200\\ 
    BIN-P            & VP + SC + SO + MI + BP    &$2$  &RGB-D &180   &600-1200\\
    T-LESS          & VP + C + TL + O + MI + SLD &$30$ &RGB-D &10080 &600-1200\\
    \bottomrule
  \end{tabular}
}
\label{tab_1} 
\end{minipage}%
\end{table*}
\section{Evaluation Metrics}
\label{Evaluation_met}
\index{evaluation metrics}
Several evaluation metrics have been proposed to determine whether an estimated 6D pose is correct. The multi-modal analyses of instance-level methods presented in Sect. \ref{Exp_Res} are based on the Average Distance (AD) metric \cite{32}, and the multi-modal analyses of category-level methods are based on 3D Intersection over Union (IoU) \cite{2}. These two metrics are detailed in this section.\\
\index{evaluation metrics!average distance}
\indent \textit{\textbf{Average Distance (AD).}} This is one of the most widely used metrics in the literature \cite{32}. Given the ground truth $(\bar{\mathbf{x}}, \bar{\mathbf{\theta}})$ and estimated $(\mathbf{x}, \mathbf{\theta})$ poses of an object of interest $O$, this metric outputs $\omega_\mathrm{AD}$, the score of the average distance between $(\bar{\mathbf{x}}, \bar{\mathbf{\theta}})$ and $(\mathbf{x}, \mathbf{\theta})$. It is calculated over all points $\mathbf{s}$ of the $3$D model $M$ of the object of interest:
\begin{equation}
\omega_\mathrm{AD} = \underset{\mathbf{s} \in M}{\mathrm{avg}} \vert \vert (\bar{R}\mathbf{s} + \bar{T}) - (R\mathbf{s} + T) \vert \vert
\label{eq5a}
\end{equation}
where $\bar{R}$ and $\bar{T}$ depict rotation and translation matrices of the ground truth pose $(\bar{\mathbf{x}}, \bar{\mathbf{\theta}})$, while $R$ and $T$ represent rotation and translation matrices of the estimated pose $(\mathbf{x}, \mathbf{\theta})$. Hypotheses ensuring the following inequality are considered as correct:
\begin{equation}
\omega_\mathrm{AD} \leq z_{\omega} \Phi
\label{eq4}
\end{equation}
where $\Phi$ is the diameter of the 3D model $M$, and $z_{\omega}$ is a constant that determines the coarseness of a hypothesis which is assigned as correct. Note that, Eq. \ref{eq5a} is valid for objects whose models are not ambiguous or do not have any subset of views under which they appear to be ambiguous. In case the model $M$ of an object of interest has indistinguishable views, Eq. \ref{eq5a} transforms into the following form:
\begin{equation}
\omega_\mathrm{AD} = \underset{\mathbf{s_1} \in M}{\mathrm{avg}}\underset{\mathbf{s_2} \in M}{\mathrm{min}} \vert \vert (\bar{R}\mathbf{s_1} + \bar{T}) - (R\mathbf{s_2} + T) \vert \vert
\label{eq5}
\end{equation}
where $\omega_\mathrm{AD}$ is calculated as the average distance to the closest model point. This function employs many-to-one point matching and significantly promotes symmetric and occluded objects, generating lower $\omega_\mathrm{AD}$ scores.\\
\index{evaluation metrics!intersection over union}
\indent \textit{\textbf{Intersection over Union.}} This metric is originally presented to evaluate the performance of the methods working in 2D space. Given the estimated and ground truth bounding boxes $B$ and $\bar{B}$ and assuming that they are aligned with image axes, it determines the area of intersection $B \cap \bar{B}$, and the area of union $B \cup \bar{B}$, and then comparing these two, outputs the overlapping ratio $\omega_\mathrm{IoU}$:
\begin{equation}
\omega_\mathrm{IoU} = \frac{B \cap \bar{B}}{B \cup \bar{B}}
\label{eq9}
\end{equation}
According to Eq. \ref{eq9}, a predicted box is considered to be correct if the overlapping ratio $\omega_\mathrm{IoU}$ is more than the threshold $\tau_\mathrm{IoU} = 0.5$. This metric is further extended to work with 3D volumes calculating overlapping ratio $\omega_\mathrm{IoU_{3D}}$ over 3D bounding boxes \cite{2}. The extended version assumes that 3D bounding boxes are aligned with gravity direction, but makes no assumption on the other two axes.\\
\indent In this study, we employ a twofold evaluation strategy for the instance-level 6D object detectors using the AD metric: i) Recall. The hypotheses on the test images of every object are ranked, and the hypothesis with the highest weight is selected as the estimated 6D pose. Recall value is calculated comparing the number of correctly estimated poses and the number of the test images of the interested object. ii) F1 scores. Unlike recall, all hypotheses are taken into account, and F1 score, the harmonic mean of precision and recall values, is presented. For evaluating the category-level detectors, the 3D IoU metric is utilized, and Average Precision (AP) results are provided.
\section{Multi-modal Analyses}
\label{Exp_Res}
\index{analysis}
At the level of instances, we analyse ten baselines on the datasets with respect to both challenges and the architectures. Two of the baselines \cite{32, 20} are our own implementations. The first implementation is of Linemod \cite{32}. Since it is one of the methods been at the forefront of 6D object pose estimation research, we choose this method for implementation to enhance our analyses on the challenges. It is based on templates, and frequently been compared by the state-of-the-art. We compute the color gradients and surface normal features using the built-in functions and classes provided by OpenCV. Our second implementation is Latent-Class Hough Forests (LCHF) \cite{20}. There are high number of learning based 6D object pose estimation methods in the literature, using random forests as regressor. We have implemented LCHF, since it representatively demonstrates the characteristics of regressors. The features in Latent-Class Hough Forest (LCHF) are the part-based version of the features introduced in \cite{32}. Hence, we inherit the classes given by OpenCV in order to generate part-based features used in LCHF. We train each method for the objects of interest by ourselves, and using the learnt classifiers, we test those on all datasets. Note that, the methods use only foreground samples during training/template generation. In this section, \enquote{LINEMOD} refers to the dataset, whilst \enquote{Linemod} is used to indicate the baseline itself.\\
\indent At the level of categories, we analyse four baselines \cite{1, 2, 4, 75}, one of which is our own implementation \cite{75}. The architecture presented in \cite{75} is a part-based random forest architecture. Its learning scheme is privileged. The challenges of the categories are learnt during training in which 3D skeleton structures are derived as shape-invariant features. In the test, there is no skeleton data, and the depth pixels are directly used as features in order to vote the 6D pose of an instance, given the category of interest.
\index{analysis!at the level of instances}
\subsection{Analyses at the Level of Instances}
Utilizing the AD metric, we compare the chosen baselines along with the challenges, i) regarding the recall values that each baseline generates on every dataset, ii) regarding the F1 scores. The coefficient $z_{\omega}$ is $0.10$, and in case we use different thresholds, we will specifically indicate in the related parts.\\

\begin{figure}[!t]
\centering
\includegraphics[width=\textwidth]{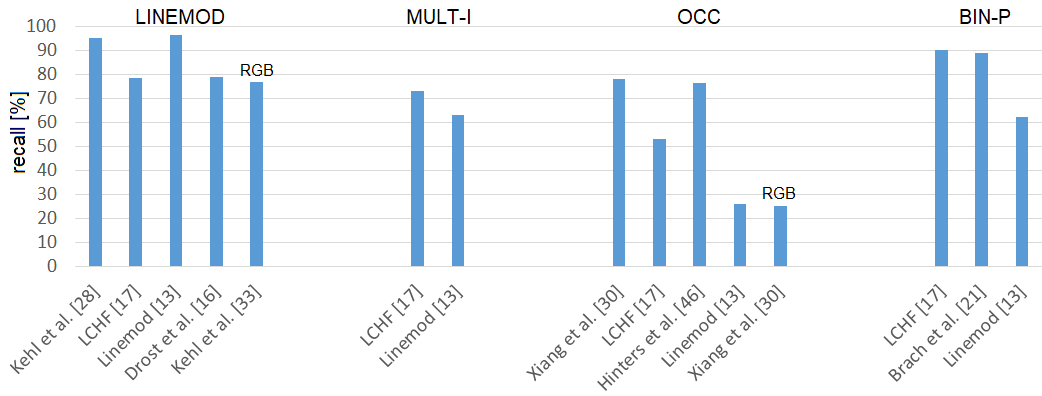}
\caption{Success of each baseline on every dataset is shown, recall values are computed using the Average Distance (AD) metric.}
\label{fig2_a}
\end{figure}

\begin{figure}[!t]
\centering
\includegraphics[width=\textwidth]{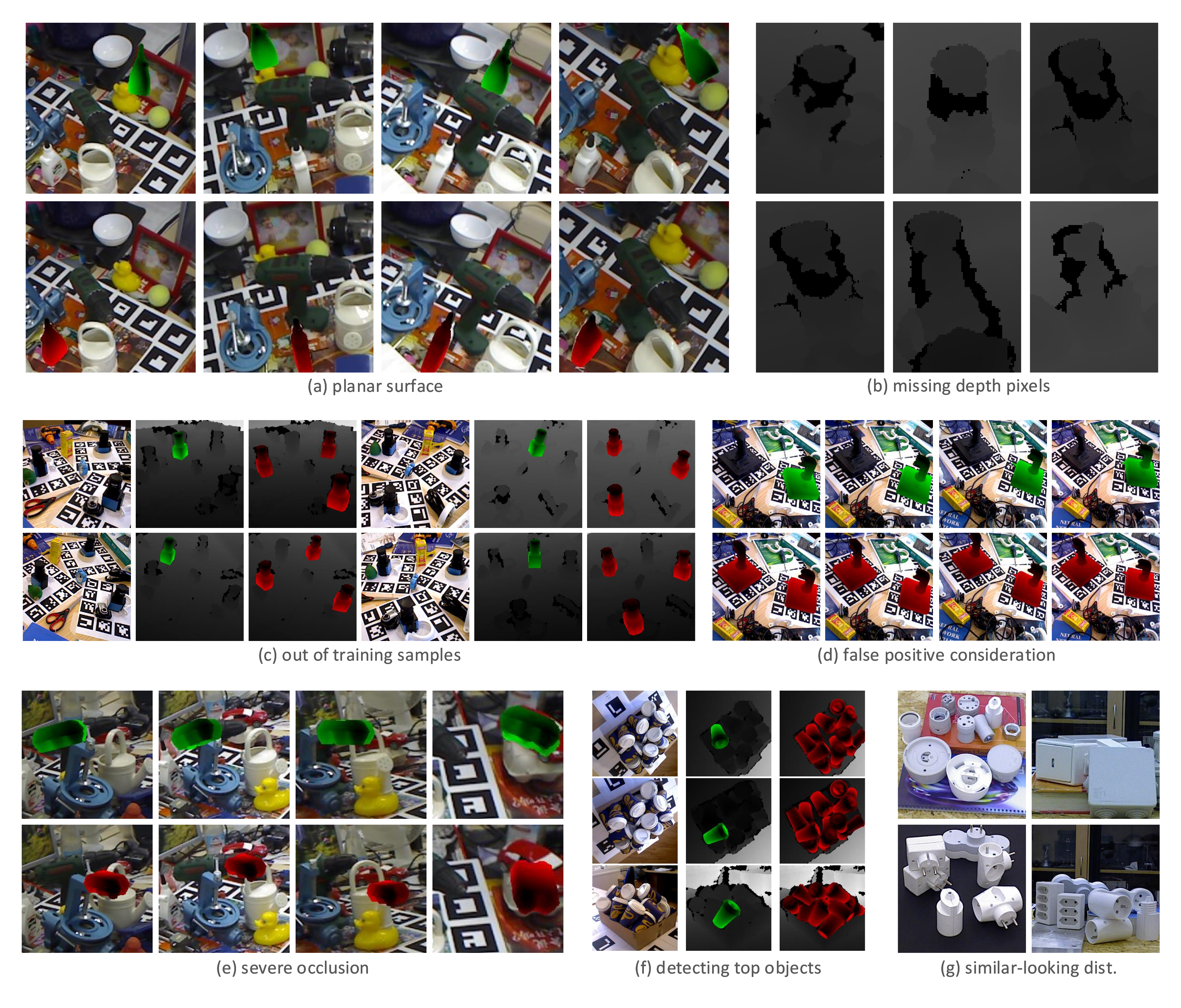}
\caption{(a)-(g) challenges encountered during test are exemplified (green renderings are hypotheses, and the red ones are ground truths).}
\label{fig2_b}
\end{figure}

\noindent \textbf{Recall-only Discussions:} Recall-only discussions are based on the numbers provided in Table \ref{tab_2}, and Fig. \ref{fig2_a}.\\
\index{challenges of instances!clutter}
\index{challenges of instances!viewpoint variability}
\index{challenges of instances!texture-less objects}
\indent \textit{\textbf{Clutter, Viewpoint, Texture-less objects.}} Highest recall values are obtained on the LINEMOD dataset (see Fig. \ref{fig2_a}), meaning that the state-of-the-art methods for 6D object pose estimation can successfully handle the challenges, clutter, varying viewpoint, and texture-less objects. LCHF, detecting more than half of the objects with over $80 \%$ accuracy, worst performs on \enquote{box} and \enquote{glue} (see Table \ref{tab_2a}), since these objects have planar surfaces, which confuses the features extracted in depth channel (example images are given in Fig. \ref{fig2_b} (a)).\\
\index{challenges of instances!occlusion}
\indent \textit{\textbf{Occlusion.}} In addition to the challenges involved in LINEMOD, occlusion is introduced in MULT-I. Linemod's performance decreases, since occlusion affects holistic feature representations in color and depth channels. LCHF performs better on this dataset than Linemod. Since LCHF is trained using the parts coming from positive training images, it can easily handle occlusion, using the information acquired from occlusion-free parts of the target objects. However, LCHF degrades on \enquote{camera}. In comparison with the other objects in the dataset, \enquote{camera} has relatively smaller dimensions. In most of the test images, there are non-negligible amount of missing depth pixels (Fig. \ref{fig2_b} (b)) along the borders of this object, and thus confusing the features extracted in depth channel. In such cases, LCHF is liable to detect similar-looking out of training objects and generate many false positives (see Fig. \ref{fig2_b} (c)). The hypotheses produced by LCHF for \enquote{joystick} are all considered as false positive (Fig. \ref{fig2_b} (d)). When we re-evaluate the recall that LCHF produces on the \enquote{joystick} object setting $z_{\omega}$ to the value of $0.15$, we observe $89 \%$ accuracy.\\
\index{challenges of instances!occlusion}
\indent \textit{\textbf{Severe Occlusion.}} OCC involves challenging test images where the objects of interest are cluttered and severely occluded. The best performance on this dataset is caught by Xiang et al. \cite{72}, and there is still room for improvement in order to fully handle this challenge. Despite the fact that the distinctive feature of this benchmark is the existence of \enquote{severe occlusion}, there are occlusion-free target objects in several test images. In case the test images of a target object include unoccluded and/or naively occluded samples (with the occlusion ratio up to $40 \% -50 \%$ of the object dimensions) in addition to severely occluded samples, methods produce relatively higher recall values (\textit{e.g.} \enquote{can, driller, duck, holepuncher}, Table \ref{tab_2c}). On the other hand, when the target object has additionally other challenges such as planar surfaces, methods' performance (LCHF and Linemod) decreases (\textit{e.g.} \enquote{box}, Fig. \ref{fig2_b} (e)).\\
\index{challenges of instances!clutter}
\indent \textit{\textbf{Severe Clutter.}} In addition to the challenges discussed above, BIN-P inherently involves severe clutter, since it is designed for bin-picking scenarios, where objects are arbitrarily stacked in a pile. According to the recall values presented in Table \ref{tab_2d}, LCHF and Brachmann et al. \cite{31} perform $ 25 \%$ better than Linemod. Despite having severely occluded target objects in this dataset, there are unoccluded/relatively less occluded objects at the top of the bin. Since our current analyses are based on the top hypothesis of each method, the produced success rates show that the methods can recognize the objects located on top of the bin with reasonable accuracy (Fig. \ref{fig2_b} (f)).\\
\index{challenges of instances!similar looking distractors}
\indent \textit{\textbf{Similar-Looking Distractors.}} We test both Linemod and LCHF on the T-LESS dataset. Since most of the time the algorithms fail, we do not report quantitative analyses, instead we discuss our observations from the experiments. The dataset involves various object classes with strong shape and color similarities. When the background color is different than that of the objects of interest, color gradient features are successfully extracted. However, the scenes involve multiple instances, multiple objects similar in shape and color, and hence, the features queried exist in the scene at multiple locations. The features extracted in depth channel are also severely affected from the lack of discriminative selection of shape information. When the objects of interest have planar surfaces, the detectors cannot easily discriminate foreground and background in depth channel, since these objects in the dataset are relatively smaller in dimension (see Fig. \ref{fig2_b} (g)).\\
\index{challenges of instances!occlusion}
\indent \textit{\textbf{Part-based vs. Holistic approaches.}} Holistic methods \cite{32, 64, 66, 72, 68} formulate the detection problem globally. Linemod \cite{32} represents the windows extracted from RGB and depth images by the surface normals and color gradients features. Distortions along the object borders arising from occlusion and clutter, that is, the distortions of the color gradient and surface normal information in the test processes, mainly degrade the performance of this detector. Part-based methods \cite{20, 31, 18, 21, 33} extract parts in the given image. Despite the fact that LCHF uses the same kinds of features as in Linemod, LCHF detects objects extracting parts, thus making the method more robust to occlusion and clutter. As illustrated in Table \ref{tab_2}, the part-based method LCHF consistently overperforms the holistic method Linemod.\\
\index{instance-based methods!template-based}
\index{instance-based methods!conventional learning-based}
\indent \textit{\textbf{Template-based vs. Random forest-based.}} Template-based methods, \textit{i.e.}, Linemod, match the features extracted during test to a set of templates, and hence, they cannot easily be generalized well to unseen ground truth annotations, that is, the translation and rotation parameters in object pose estimation. Methods based on random forests \cite{20, 31, 18, 21} efficiently benefit the randomisation embedded in this learning tool, consequently providing good generalisation performance on new unseen samples. Table \ref{tab_2} clearly depicts that methods based on random forests \cite{20, 31, 18, 21} generate higher recall values than template-based Linemod.\\
\index{instance-based methods!point-to-point}
\indent \textit{\textbf{RGB-D vs. Depth.}} Methods utilizing both RGB and depth channels demonstrate higher recall values than methods that are of using only depth, since RGB provides extra clues to ease the detection. This is depicted in Table \ref{tab_2a} where learning- and template-based methods of RGB-D perform much better than point-to-point technique \cite{64} of depth channel.\\
\index{instance-based methods!deep learning}
\indent \textit{\textbf{RGB-D vs. RGB (CNN structures).}} More recent paradigm is to adopt CNNs to solve 6D object pose estimation problem taking RGB images as inputs. In Table \ref{tab_2}, the methods \cite{72, 68} are based on CNN structure. According to Table \ref{tab_2a}, SSD-6D, the deep method of Kehl et al. \cite{69}, produces $76.7 \%$ recall value. Despite the fact that it shows the minimum performance on the LINEMOD dataset, it is important to consider that the method is trained and tested in RGB channel, while the remaining methods additionally use the depth data. The method of Xiang et al. \cite{72} is evaluated on OCC dataset in both RGB-D and RGB channel. The best performance on the OCC dataset is demonstrated by the deep method of Xiang et al. \cite{72}, in case it is trained and is evaluated in RGB-D channel. However, its performance degrades when trained only using RGB data.\\
\begin{table*}[t]
\caption{\label{tab_2}Methods' performance are depicted object-wise based on recall values computed using the Average Distance (AD) evaluation protocol.}
\vspace{0.8em}
\centering

\begin{minipage}{0.99\textwidth}
\centering
\setlength\tabcolsep{3pt}
\resizebox{0.99\columnwidth}{!}{
\begin{tabular}[t]{c c c c c c c c c c c c c c c c}
    \toprule
  \textbf{Method}       &ch.     &ape    &bvise & cam & can & cat & dril & duck & box & glue & hpunch & iron & lamp & phone & \textbf{AVER}\\  
  \midrule
  Kehl et al \cite{33}  &RGB-D   &96.9   &94.1  &97.7 &95.2 &97.4 &96.2 &97.3 &99.9 &78.6 &96.8 &98.7 &96.2 &92.8 &95.2\\
  LCHF \cite{20}        &RGB-D   &84	 &95    &72   &74 &91 &92 &91 &48 &55 &89 &72 &90 &69 &78.6\\
  Linemod \cite{32}     &RGB-D   &95.8   &98.7  &97.5 &95.4 &99.3 &93.6 &95.9 &99.8 &91.8 &95.9 &97.5 &97.7 &93.3 &96.3\\
  Drost et al \cite{64} &D       &86.5   &70.7  &78.6 &80.2 &85.4 &87.3 &46 &97 &57.2 &77.4 &84.9 &93.3 &80.7  &78.9\\
  Kehl et al \cite{68}  &RGB     &65     &80    &78   &86 &70 &73 &66 &100  &100  &49 &78 &73 &79 &76.7\\
\bottomrule
  \end{tabular}
}
\vspace{-1.0em}
\caption*{(a) LINEMOD dataset}
\vspace{0.9em}
\label{tab_2a} 
\end{minipage}%

\begin{minipage}{0.80\textwidth}
\centering
\setlength\tabcolsep{6pt}
\resizebox{0.90\columnwidth}{!}{
\begin{tabular}[t]{c c c c c c c c c}
\toprule
\textbf{Method}     &ch.   &camera  &cup  &joystick   &juice   &milk   &shampoo & \textbf{AVER}\\
\midrule
 LCHF \cite{20}     &RGB-D &52.5    &99.8 &0       &99.3    &92.7   &97.2 &73\\
 Linemod \cite{32}  &RGB-D &18.3    &99.2 &85         &51.6    &72.2   &53.1 &63.2\\
\bottomrule
\end{tabular}
}
\vspace{-1.0em}
\caption*{(b) MULT-I dataset}
\vspace{0.9em}
\label{tab_2b} 
\end{minipage}%

\begin{minipage}{0.80\textwidth}
\centering
\setlength\tabcolsep{6pt}
\resizebox{0.90\columnwidth}{!}{
\begin{tabular}[t]{c c c c c c c c c c c}
\toprule
\textbf{Method} &ch. &ape &can &cat &drill &duck &box &glue &hpunch &\textbf{AVER}\\ 
\midrule
    Xiang et al. \cite{72}    &RGB-D  &76.2	 &87.4  &52.2 &90.3 &77.7 &72.2   &76.7   &91.4  &78\\
    LCHF \cite{20}            &RGB-D  &48.0	 &79.0  &38.0 &83.0 &64.0 &11.0   &32.0   &69.0  &53\\
    Hinters et al. \cite{66}  &RGB-D  &81.4	 &94.7  &55.2 &86.0 &79.7 &65.5   &52.1   &95.5  &76.3\\
    Linemod \cite{32}         &RGB-D  &21.0	 &31.0  &14.0 &37.0 &42.0 &21.0   &5.0    &35.0  &25.8\\
    Xiang et al. \cite{72}    &RGB    &9.6	 &45.2  &0.93 &41.4 &19.6 &22.0   &38.5   &22.1  &25\\
\bottomrule
\end{tabular}
}
\vspace{-1.0em}
\caption*{(c) OCC dataset}
\vspace{0.9em}
\label{tab_2c} 
\end{minipage}%

\begin{minipage}{0.60\textwidth}
\centering
\setlength\tabcolsep{12pt}
\resizebox{0.90\columnwidth}{!}{
\begin{tabular}[t]{c c c c c}
\toprule
\textbf{Method}                &ch.      &cup         &juice  &\textbf{AVER}\\
\midrule
	LCHF \cite{20}   &RGB-D    &90.0	      &89.0   &90\\
	Brach et al. \cite{31}   &RGB-D    &89.4        &87.6   &89\\
	Linemod \cite{32}   &RGB-D    &88.0	      &40.0   &64\\
\bottomrule
\end{tabular}
}
\vspace{-1.0em}
\caption*{(d) BIN-P dataset}
\vspace{0.9em}
\label{tab_2d} 
\end{minipage}%
\end{table*}
\begin{table*}[t]
\caption{\label{tab_3} Methods' performance are depicted object-wise based on F1 scores computed using the Average Distance (AD) evaluation protocol.}
\vspace{0.8em}
\centering
\setlength\tabcolsep{6pt}

\begin{minipage}{0.99\textwidth}
\centering
\resizebox{0.99\columnwidth}{!}{
\begin{tabular}[t]{c c c c c c c c c c c c c c c c}
\toprule
\textbf{Method} &ch. & ape & bvise & cam & can & cat & dril & duck & box & glue & hpunch & iron & lamp & phone & \textbf{AVER}\\ 
\midrule
    Kehl et al. \cite{33}  &RGB-D &0.98 &0.95 &0.93 &0.83 &0.98 &0.97 &0.98 &1 &0.74 &0.98 &0.91 &0.98 &0.85 &0.93\\
    LCHF \cite{20}  &RGB-D &0.86 &0.96 &0.72 &0.71 &0.89 &0.91 &0.91 &0.74 &0.68 &0.88 &0.74 &0.92 &0.73 &0.82\\
    Linemod \cite{32} &RGB-D  &0.53 &0.85 &0.64 &0.51 &0.66 &0.69 &0.58 &0.86 &0.44 &0.52 &0.68 &0.68 &0.56 &0.63\\
    Kehl et al. \cite{68}  &RGB &0.76 &0.97 &0.92 &0.93 &0.89 &0.97 &0.80 &0.94 &0.76 &0.72 &0.98 &0.93 &0.92 &0.88\\
\bottomrule
  \end{tabular}
}
\vspace{-1.0em}
\caption*{(a) LINEMOD dataset}
\vspace{0.9em}
\label{tab_3a} 
\end{minipage}%

\begin{minipage}{0.90\textwidth}
\centering
\resizebox{0.99\columnwidth}{!}{
\begin{tabular}[t]{c c c c c c c c c}
\toprule
\textbf{Method}      &ch.         &camera  &cup  &joystick  &juice  &milk  &shampoo & \textbf{AVER}\\
\midrule
Kehl et al. \cite{33}     &RGB-D     &0.38  &0.97  &0.89	 &0.87  &0.46  &0.91 &0.75\\
LCHF \cite{20}   &RGB-D     &0.39  &0.89  &0.55	 &0.88  &0.40  &0.79 &0.65\\
Drost et al. \cite{64}    &D     &0.41  &0.87  &0.28	 &0.60  &0.26  &0.65 &0.51\\
Linemod \cite{32}  &RGB-D     &0.37  &0.58  &0.15     &0.44  &0.49  &0.55 &0.43\\	
Kehl et al. \cite{68}     &RGB     &0.74  &0.98  &0.99  &0.92  &0.78  &0.89 &0.88\\
\bottomrule
\end{tabular}
}
\vspace{-1.0em}
\caption*{(b) MULT-I dataset}
\vspace{0.9em}
\label{tab_3b} 
\end{minipage}%

\begin{minipage}{0.80\textwidth}
\centering
\resizebox{0.99\columnwidth}{!}{
\begin{tabular}[t]{c c c c c c c c c c c}
\toprule
\textbf{Method} &ch. &ape &can &cat &dril &duck &box &glue &hpunch & \textbf{AVER}\\ 
\midrule
   LCHF \cite{20}  &RGB-D  &0.51	 &0.77  &0.44 &0.82 &0.66 &0.13 &0.25 &0.64 &0.53\\
   Linemod \cite{32}  &RGB-D  &0.23	 &0.31  &0.17 &0.37 &0.43 &0.19 &0.05 &0.30 &0.26\\
  Brach et al. \cite{21}   &RGB  &- &- &- &- &- &- &- &- &0.51\\
   Kehl et al. \cite{68}   &RGB  &- &- &- &- &- &- &- &- &0.38\\
\bottomrule
\end{tabular}
}
\vspace{-1.0em}
\caption*{(c) OCC dataset}
\vspace{0.9em}
\label{tab_3c} 
\end{minipage}%

\begin{minipage}{0.70\textwidth}
\centering
\setlength\tabcolsep{12pt}
\resizebox{0.99\columnwidth}{!}{
\begin{tabular}[t]{c c c c c}
\toprule
\textbf{Method}              &ch.    &cup   &juice &\textbf{AVER}\\
\midrule
           LCHF \cite{20}    &RGB-D  &0.48  &0.29  &0.39\\
    Doumanoglou et al. \cite{18}    &RGB-D  &0.36  &0.29  &0.33\\
        Linemod \cite{32}    &RGB-D  &0.48  &0.20  &0.34\\
\bottomrule
\end{tabular}
}
\vspace{-1.4em}
\caption*{(d) BIN-P dataset}
\vspace{0.9em}
\label{tab_3d} 
\end{minipage}%
\end{table*}
\indent Robotic manipulators that pick and place the items from conveyors, shelves, pallets, \textit{etc.}, need to know the pose of one item per RGB-D image, even though there might be multiple items in its workspace. Hence our recall-only analyses mainly target to solve the problems that could be encountered in such cases. Based upon the analyses currently made, one can make important implications, particularly from the point of the performances of the detectors. On the other hand, recall-based analyses are not enough to illustrate which dataset is more challenging than the others. This is especially true in crowded scenarios where multiple instances of target objects are severely occluded and cluttered. Therefore, in the next part, we discuss the performances of the baselines from another aspect, regarding precision-recall curves and F1 scores, where the 6D detectors are investigated sorting all detection scores across all images.\\

\begin{figure*}[!t]
\captionsetup[subfigure]{labelformat=empty}
\centering
\includegraphics[width=\linewidth]{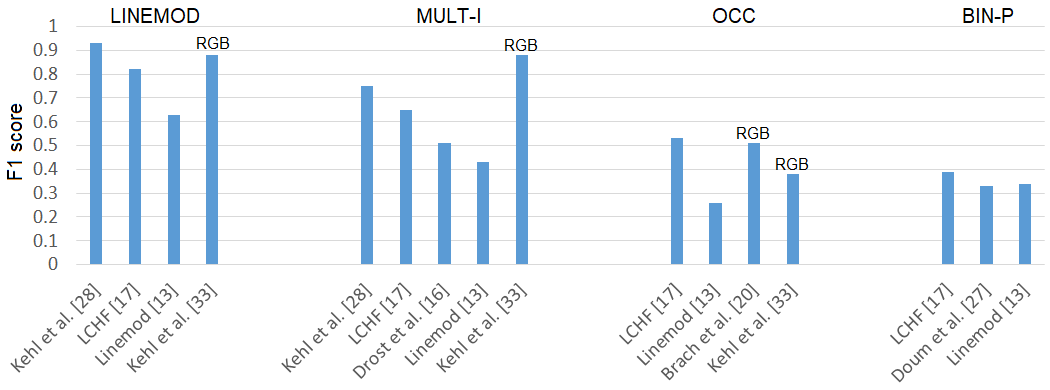}
\caption{Success of each baseline on every dataset is shown, F1 scores are computed using the Average Distance (AD) metric.}
\label{fig3_1}
\end{figure*}

\noindent \textbf{Precision-Recall Discussions:} Our precision-recall discussions are based on the F1 scores provided in Table \ref{tab_3}, and Fig. \ref{fig3_1}.\\
\indent We first analyse the methods \cite{32, 20, 68, 33} on the LINEMOD dataset. On the average, Kehl et al. \cite{33} outperforms other methods proving the superiority of learning deep features. Despite estimating 6D in RGB images, SSD-6D \cite{68} exhibits the advantages of using CNN structures for 6D object pose estimation. LCHF and Linemod demonstrate lower performance, since the features used by these methods are manually-crafted. The comparison between Fig. \ref{fig2_a} and Fig. \ref{fig3_1} reveals that the results produced by the methods have approximately the same characteristics on the LINEMOD dataset, with respect to recall and F1 scores.\\
\begin{figure*}[!t]
\captionsetup[subfigure]{labelformat=empty}
\centering
\includegraphics[width=\linewidth]{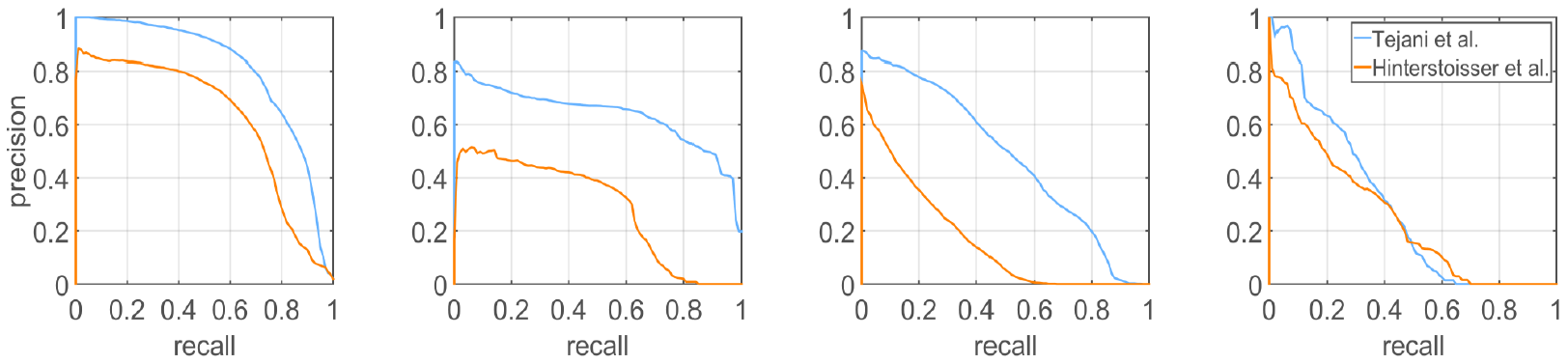}
\caption{Precision-recall curves of averaged F1 scores for Tejani et al. \cite{20} and Hinterstoisser et al. \cite{32} are shown: from left to right, LINEMOD, MULT-I, OCC, BIN-P.}
\label{fig3_2}
\end{figure*}
\indent The methods tested on the MULT-I dataset \cite{33, 20, 64, 32, 68} utilize the geometry information inherently provided by depth images. Despite this fact, SSD-6D \cite{68}, estimating 6D pose only from RGB images, outperforms other methods clearly proving the superiority of using CNNs for the 6D problem over other structures.\\
\indent LCHF \cite{20} and Brachmann et al. \cite{21} best perform on OCC with respect to F1 scores. As this dataset involves test images where highly occluded objects are located, the reported results depict the importance of designing part-based solutions.\\
\indent The most important difference is observed on the BIN-P dataset. While the success rates of the detectors on this dataset are higher than $60 \%$ with respect to the recall values (see Fig. \ref{fig2_a}), according to the presented F1 scores, their performance are less than $40 \%$. When we take into account all hypotheses and the challenges particular to this dataset, which are severe occlusion and severe clutter, we observe strong degradation in the accuracy of the detectors.\\
\indent In Fig. \ref{fig3_2}, we lastly report precision-recall curves of LCHF and Linemod. Regarding these curves, one can observe that as the datasets are getting more difficult, from the point of challenges involved, the methods produce less accurate results.\\
\index{analysis!at the level of categories}
\index{challenges of instances!clutter}
\index{challenges of instances!occlusion}
\index{category-based methods!3D}
\index{category-based methods!4D}
\subsection{Analyses at the Level of Categories}
\begin{figure}[!t]
\captionsetup[subfigure]{labelformat=empty}
\centering
\includegraphics[width=\linewidth]{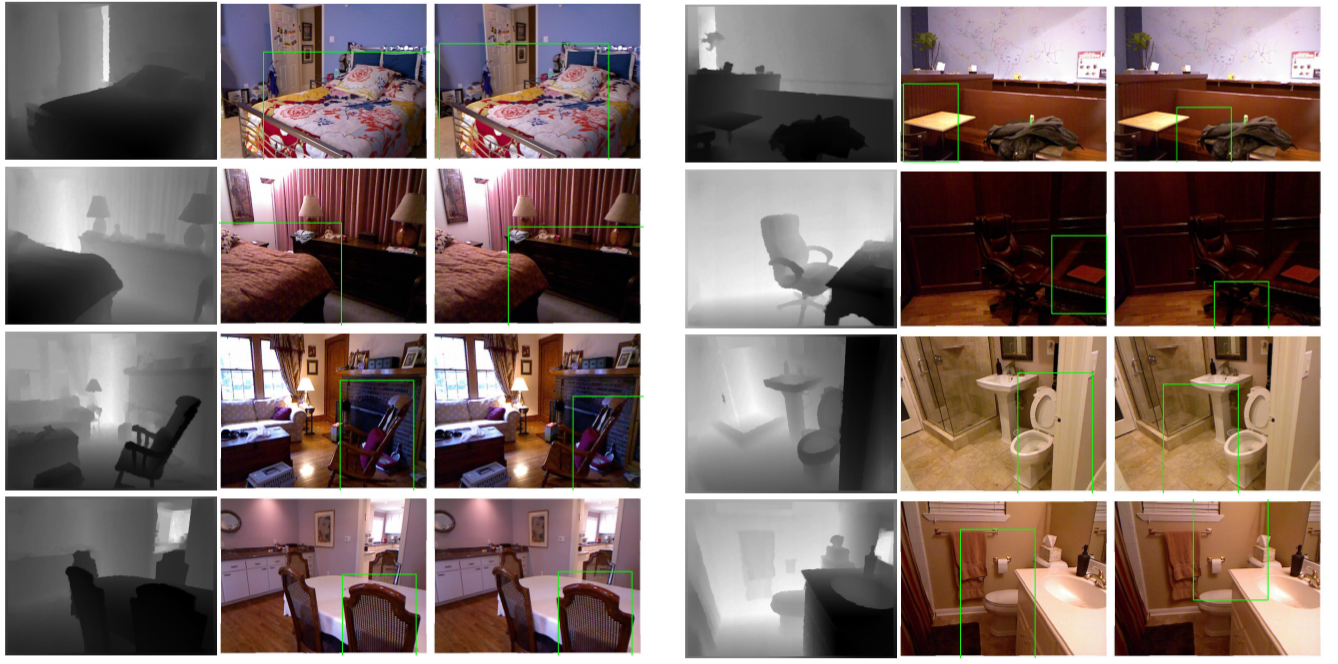}
\caption{Sample results generated by ISA on real data: (for each triplet) each row is for per scene. First column depicts depth images of scenes. Estimations in the middle belong to ISAs trained using $Q_1\&Q_2\&Q_3$ (using 3D skeleton representation as privileged data), and hypotheses on the right are of ISAs trained on $Q_1$ only (no 3D skeleton information).}
\label{ch5_fig7_8}
\end{figure}

\begin{figure}[!t]
\captionsetup[subfigure]{labelformat=empty}
\centering
\includegraphics[width=\linewidth]{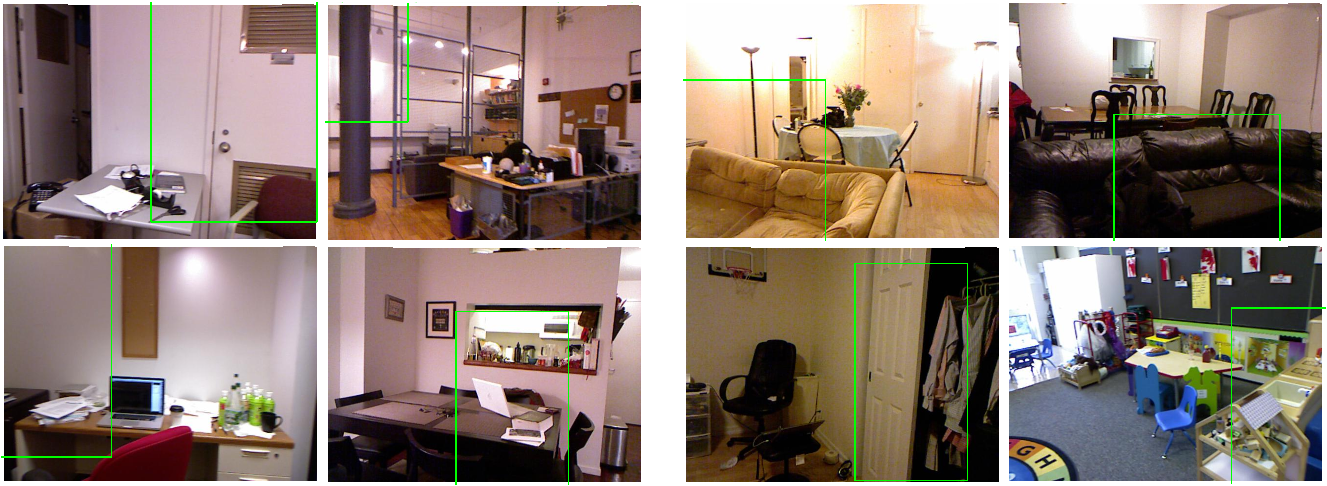}
\caption{Sample unsuccessful results generated by ISA on real data: (first row) the category of interest is table. (second row) the interested category is chair. All hypotheses are of ISAs trained using $Q_1\&Q_2\&Q_3$ (using 3D skeleton representation as privileged data).}
\label{ch5_fig7_8_unsuccess}
\end{figure}

\indent Our analyses at the level of categories are based on Table \ref{ch5_tab:result_right} and Fig. \ref{ch5_fig7_8}. Table \ref{ch5_tab:result_right} depicts the test results of the methods \cite{75, 2, 1, 4} on the RMRC dataset \cite{x13} evaluated using the metric in \cite{2}. A short analysis on the table reveals ISA demonstrate $50 \%$ average precision. The highest value ISA reaches is on the \textit{toilet} category, mainly because of the limited deviation in shape in between the instances. ISA next best performs on \textit{bed}, with $52 \%$ mean precision. The accuracy on both the categories \textit{bed} and \textit{table} are approximately the same. Despite the fact that all forests used in the experiments undergo relatively a naive training process, the highest number of the instances during training are used for the chair category. However, ISA worst performs on this category, since the images in the test dataset have strong challenges of the instances, such as occlusion, clutter, and high diversity from the shape point of view. On the average, Deep Sliding Shapes \cite{1} outperform other methods including ISA. In the real experiments, the utilization of the forest trained on less number of data and ground truth information degrades the performance of ISA across the problem's challenges. Sample results are lastly presented in Fig. \ref{ch5_fig7_8}. In these figures, the leftmost images are the inputs of ISA, and the $2^{nd}$ and the $3^{rd}$ columns demonstrate the estimations of the forests, whose training are based on $Q_1\&Q_2\&Q_3$ and $Q_1$, respectively. Training the forest using the quality function $Q_1\&Q_2\&Q_3$ stands for utilizing 3D skeleton representation as privileged data for training. However, in case the forest is trained using the quality function $Q_1$, ISA does not use any skeleton information during training For further details, check the architecture of ISA \cite{75}. Figure \ref{ch5_fig7_8_unsuccess} demonstrates several failed results. Since ISA is designed to work in depth channel, planar surfaces confuses the features extracted in the test images with the features learnt during training, consequently resulting unsuccessful estimations.\\
\begin{table*}[t]
\caption{\label{ch5_tab:result_right} 3D object detection comparison on the RMRC dataset \cite{x13} using the evaluation metric in \cite{2}.}
\centering
\setlength\tabcolsep{8pt}
\begin{minipage}{0.99\textwidth}
\centering
\resizebox{0.99\columnwidth}{!}{
    \begin{tabular}{ l c c c c c c}
    \toprule
    \textbf{Method}                     &input channel &bed           &chair         &table         &toilet &mean     \\
    \midrule    
    Sliding Shapes \cite{2}             &depth         &33.5          &29            &34.5          &67.3   &41.075    \\
    \cite{4} on instance seg.           &depth         &71            &18.2          &30.4          &63.4   &45.75    \\
    \cite{4} on estimated model         &depth         &72.7          &47.5          &40.6          &72.7   &58.375    \\
    Deep Sliding Shapes \cite{1}        &depth         &83.0          &58.8          &68.6          &79.2   &72.40    \\
    ISA \cite{75}    &depth         &52.0          &36.0          &46.5          &67.7   &50.55   \\
    \bottomrule

  \end{tabular}
  }
\end{minipage}%
\end{table*}
\section{Discussions and Conclusions}
We outline our key observations that provide guidance for future research.\\
\index{challenges of instances!clutter}
\index{challenges of instances!occlusion}
\index{challenges of instances!similar looking distractors}
\indent At the level of instances, from the challenges aspect, reasonably accurate results have been obtained on textured-objects at varying viewpoints with cluttered backgrounds. In case occlusion is introduced in the test scenes, depending on the architecture of the baseline, good performance demonstrated. Part-based solutions can handle the occlusion problem better than the ones global, using the information acquired from occlusion-free parts of the target objects. However, heavy existence of occlusion and clutter severely affects the detectors. It is possible that modelling occlusion during training can improve the performance of a detector across severe occlusion. But when occlusion is modelled, the baseline could be data-dependent. In order to maintain the generalization capability of the baseline contextual information can additionally be utilized during the modelling. Currently, similar looking distractors along with similar looking object classes seem the biggest challenge in recovering instances' 6D, since the lack of discriminative selection of shape features strongly confuse recognition systems. One possible solution could be considering the instances that have strong similarity in shape in a same category. In such a case, detectors trained using the data coming from the instances involved in the same category can report better detection results.\\
\index{instance-based methods!template-based}
\indent Architecture-wise, template-based methods, matching model features to the scene, and random forest based learning algorithms, along with their good generalization performance across unseen samples, underlie object detection and 6D pose estimation. Recent paradigm in the community is to learn deep discriminative feature representations. Despite the fact that several methods addressed 6D pose estimation utilizing deep features \cite{18, 33}, end-to-end neural network-based solutions for 6D object pose recovery are still not widespread. Depending on the availability of large-scale 6D annotated depth datasets, feature representations can be learnt on these datasets, and then the learnt representations can be customized for the 6D problem.\\
\indent These implications are related to automation in robotic systems. The implications can provide guidance for robotic manipulators that pick and place the items from conveyors, shelves, pallets, \textit{etc}. Accurately detecting objects and estimating their fine pose under uncontrolled conditions improves the grasping capability of the manipulators. Beyond accuracy, the baselines are expected to show real-time performance. Although the detectors we have tested cannot perform real-time, their run-time can be improved by utilizing APIs like OpenMP.\\
\index{challenges of categories!intra-class variation}
\index{challenges of categories!distribution shift}
\indent At the level of categories, DPMs \cite{a49, a50, a51, a53}, being at the forefront of the category-level detection research, mainly present RGB-based discrete solutions in 2D. Several studies \cite{27, 28} combine depth data with RGB. Although promising, they are not capable enough for the applications beyond 2D. More recent methods working at the level of categories are engineered to work in 3D \cite{x5, x8, x11, x12} and 4D \cite{2, 1, 4}. The ways the methods \cite{x5, x8, x11, x12, 2, 1, 4} address the challenges of categories are relatively naive. They rely on the availability of large scale 3D models in order to cover the shape variance of objects in the real world. Unlike the 3D/4D methods, ISA \cite{75} is a dedicated architecture that directly tackles the challenges of the categories, intra-class variations, distribution shifts, while estimating objects' 6D. Despite the fact that ISA is presented for category-level full 6D object pose estimation, it is tested on the real datasets, where full 6D ground truth pose annotation is unavailable. The generation of new datasets that have full 6D object pose annotations leads up the production of new methods for the category-level 6D object pose estimation problem.
\clearpage


\end{document}